\documentclass[11pt,a4paper]{article}
\usepackage{times,latexsym}
\usepackage{url}
\usepackage{booktabs}
\usepackage[T1]{fontenc}

\usepackage[final]{acl}

\usepackage{graphicx}
\usepackage{amsmath}
\usepackage{float}
\usepackage{xspace,mfirstuc,tabulary}

\usepackage{inconsolata}

\title{Compact Language, Complex Model Shifts: How and Where Ambiguity and Underspecification Affect LLMs}

\author{
  \mdseries
  \begin{tabular}{@{}c@{\hskip 2.5em}c@{}}
    \bfseries Michaela Regneri &
    \bfseries Nina Scheller \hspace{5em}  \hspace{0.2em} Sören Laue \\
    HAW Hamburg &
    Universität Hamburg \\
    Hamburg, Germany &
    Hamburg, Germany \\
    {\texttt{michaela.regneri@haw-hamburg.de}} &
    {\texttt{\{firstname.lastname\}@uni-hamburg.de}} \\
  \end{tabular}
}

\begin{document}

\maketitle
\begin{abstract}
We analyze how lexical ambiguity and underspecification affect language model training. We create artificial homonyms and artificial hypernyms as pseudowords and analyze the generative performance of language models as they are trained with increasing amounts of these ambiguous or underspecified pseudoword types. We further analyze whether the models disambiguate ambiguous or underspecified statements and provide a first mechanistic account of how ambiguity and disambiguation are represented internally. Our main results show that both ambiguity and underspecification increase model performance in ways that scale with their influence on the language's type-token ratio. However, the accuracy of generating sequences containing ambiguous words or their synonyms decreases compared to other texts. We also show that internal representations of pseudowords reflect disambiguation of pseudo-homonyms, but underspecification of pseudo-hypernyms is maintained during the generative process. 
\end{abstract}

\section{Introduction}
Ambiguity and related phenomena like underspecification are fundamental mechanisms of natural language. These phenomena have always been a main strand of research in theoretical linguistics \cite{Pinkal:1996_4}, language philosophy \cite{792f2d10-c65a-376e-b38a-492fd422a07d}, and computational linguistics \cite{hobbs-shieber-1987-algorithm,10.1093/jos/10.2.123}. Ambiguity makes language more efficient, because it allows a smaller vocabulary to express a wider range of meanings \cite[cf. the optimality theory of language,][]{PIANTADOSI2012280}. The very same mechanism then also makes language more complex, because ambiguity or underspecification has to be resolved at the levels of words, sentences, and whole discourse segments.

While much of the research on computational language understanding focused on modeling and resolving different kinds of ambiguity, the advent of large language models (LLMs) changed this discussion rapidly. Transformer models \cite{NIPS2017_3f5ee243}, the most common architecture for modern LLMs, were specifically designed to handle ambiguity through the attention mechanism. This mechanism applies context to resolve token-level ambiguities and does so hierarchically across increasingly large units of text. The impressive performance of such models seems to confirm this promise: good performance in language understanding and generation requires effective ambiguity handling, and state-of-the-art LLMs seem to handle many such cases successfully. However, the question of whether and how LLMs model ambiguity successfully is still a subject of active research, and a consequential one. Recent studies \cite{liu-etal-2023-afraid,gehring-roth-2025-ambistory} have shown that LLMs actually do fall short in handling ambiguity, even if there is evidence that their parameters encode it \cite{karamolegkou-etal-2025-trick}. 

Understanding such mechanisms matters for several reasons: On one hand, it is relevant for mitigating and avoiding \emph{bias and discrimination} in LLMs: if, e.g., a word is underspecified with respect to gender or nationality, a fair model needs to be capable of handling rare readings as well as frequent ones. On the other hand, ambiguity and underspecification make language more complex to process, while they also have the potential to compress it (as for human language processing). Given that LLMs require vast amounts of data and energy \cite{strubell-etal-2019-energy}, any additional complexity in language might make them \emph{less sustainable} if that complexity also requires more powerful models. On the other hand, language compression while maintaining performance would be utterly desirable for sustainable models. Taken together, ambiguity and underspecification as potential contributors to the cost and societal impact of LLMs should be analyzed thoroughly, and models need to process them reliably and efficiently.

Our research aims to answer whether ambiguity and underspecification in language add complexity to language model training, and how this interacts with potential effects of language compression. To assess this, we train multiple small transformer models on corpora that we modify by increasing the number of artificial homonyms and hypernyms. 
We then measure the models' performance both overall and on the newly introduced ambiguous or underspecified words. We find that, in general, neither ambiguity nor underspecification harms the models' performance. Our experiments also distinguish between setups in which we either delete synonyms of the newly introduced pseudowords or, as in natural language, maintain them, and show that the presence of synonyms harms performance, for generating both ambiguous words and their synonyms. We also give a first account showing localization of ambiguity in the model parameters, demonstrating an increased entropy in the models' MLP layers when ambiguity or underspecification increases. 
In a second step, we evaluate whether the models internally disambiguate artificial homonyms and whether they resolve the underspecification of artificial hypernyms. Our experiments show that while ambiguous terms are disambiguated in the respective context, abstract terms remain underspecified. Overall, our results can inform further research that looks at model complexity, but also on bias and model mistakes with disambiguation.

Our main contributions are as follows:
\begin{itemize}
\item We analyze the influence of ambiguity on the performance of transformer models, distinguishing between ambiguous and underspecified words. We show that both greater ambiguity and greater abstraction increase model performance linearly with the decrease in type-token ratio. We also localize the increased ambiguity and underspecification in the models' MLP layers.
\item We analyze the models' specific capability of generating ambiguous or abstract words, and show that performance on generating ambiguous expressions decreases with more ambiguity, while abstract expressions can be generated more stably.
\item By activation analysis, we show that ambiguity and underspecification differ markedly in terms of resolution: While artificial homonyms are disambiguated in the model, artificial hypernyms retain their underspecified representations.
\end{itemize}
To the best of our knowledge, this is the first study to relate ambiguity and abstraction, as sources of linguistic complexity, to language model performance, and to show how model training can benefit or suffer from them. We provide code and data for all our experiments.\footnote{\url{https://github.com/mregneri/ambiguity_abstraction_llms/}}

\section{Background and Related Work}\label{sec:related}
\subsection{Ambiguity and Underspecification}
We analyze how ambiguity and underspecification affect language model performance. Concretely, we create corpora with artificial homonyms, train language models on them, and evaluate their performance as a function of the number of homonyms. We conduct analogous experiments using artificial hypernyms and compare the results. This setup draws on and contributes to several lines of research in linguistics and machine learning.
\paragraph{Ambiguity in LLMs} has been studied extensively, particularly at the word level. A first line of work investigates how models represent ambiguity internally. \citet{https://doi.org/10.1111/cogs.12943} studied the differences between homonyms and polysemes in static word embeddings. We adapt some of their homonymy measures to filter and modify our corpora when introducing artificial homonyms. \citet{haber-poesio-2021-patterns-polysemy,haber-poesio-2024-polysemy} extended the analysis of homonyms and polysemes to LLMs, showing that their difference is predictable from the models' embedding layers. Other work has found clear evidence that ambiguity is encoded in the models' parameters \cite{park-kim-2025-llms}, and models can also be tuned to handle ambiguity explicitly with alignment algorithms \cite{kim-etal-2024-aligning}, supporting the hypothesis that they can encode ambiguity in principle.
A second line of work evaluates how models perform on ambiguity-related tasks. LLMs perform well on many tasks related to standard word sense disambiguation \cite{sumanathilaka-etal-2024-llms}, but still fall short in particularly challenging settings \cite{gehring-roth-2025-ambistory}. \citet{liu-etal-2023-afraid} provide a dataset with various types of ambiguity, and their experiments show how ambiguity affects entailment handling in LLMs, concluding that ambiguity is not properly modeled. Even when ambiguity is encoded in the parameters, models do not perform well when prompted to use it \cite{karamolegkou-etal-2025-trick}, and they often fall short in standard settings even after alignment tuning \cite{kim-etal-2024-aligning}. This gap might be due to LLMs failing to recognize relevant ambiguity in user prompts \citep{zhang-etal-2024-clamber}.

\paragraph{Underspecification in LLMs} is, in our case, operationalized as taxonomic abstraction through hypernymy at the word level. Research on abstraction in LLMs is comparatively sparser than on ambiguity. Some studies focus on performance, e.g., showing the limits of LLMs' conceptual knowledge \cite{peng-etal-2022-copen} and demonstrating that incorporating taxonomic relationships improves LLM reasoning \cite{TORRESMORENO2026114825}. Models can also be prompted to produce taxonomic hierarchies, but the concept hierarchies encoded in earlier models like BERT seem to differ from human ones \cite{dalvi2020discovering}. As with ambiguity, taxonomic abstraction is recoverable to some degree from LLM parameters \cite{regneri-etal-2024-detecting}. Beyond lexical abstraction, there are also studies on LLMs and scope underspecification \cite{wildenburg-etal-2024-pre}, which show that LLMs can detect underspecification and partially resolve it.
\paragraph{Pseudowords in LLMs} have been used to assess varying linguistic phenomena, and we use them to study the effect of ambiguity and abstraction on LLMs. Pseudowords were originally invented to create datasets for word sense disambiguation \cite{schutze-1998-automatic}. A pseudoword is an artificial term created by picking a word pair from a corpus, creating a new word similar to both original words, and then replacing occurrences of the original words with the new pseudoword.
While annotated datasets for word sense disambiguation exist by now, pseudowords are still used, e.g., to study language change \cite{shoemark-etal-2019-room} and to show how word sense disambiguation is reflected in LLMs \cite{karidi-etal-2021-putting}.
Crucially for our setup, even unknown pseudowords are processable by LLMs \cite{10.1162/coli_a_00527}. Using arbitrary strings like "wd134x" would introduce out-of-distribution tokens bearing no subword-level similarity to any natural word, potentially creating unwanted confounds. Pseudowords avoid this problem because they are composed of familiar subword tokens.

\section{Dataset Creation} \label{sec:data}
The following describes how we create our datasets for model training to measure the effects of ambiguity and underspecification.
 \subsection{Homonyms and Hypernyms} \label{sec:hypo}
We investigate word-level ambiguity and underspecification, concretely through homonyms and hypernyms. Homonyms are ambiguous words that have two disjoint readings (\emph{bat}, \emph{bank}), as opposed to polysemes, whose readings have semantic overlap to varying degrees (e.g., \emph{crane}, \emph{paper}). A hypernym is a word whose meaning encompasses that of more specific words, its hyponyms (\emph{animal} (hypernym) -- \emph{dog} (hyponym)), and can have multiple hyponyms.
Comparing homonyms with hypernyms lets us contrast the effects of ambiguous words with semantically divergent readings against words with varying levels of underspecification. In vector space terms, this means we compare words that have close neighbors in very different regions of the space (cf. Sec.~\ref{sec:space}).


\subsection{Corpora}\label{sec:corpora}
To study the effects of ambiguity and underspecification, we chose to train \emph{small} language models on \emph{narrow-domain} language corpora. Our objectives were twofold: to isolate the effects of ambiguity and abstraction by starting with corpora that contain as little ambiguity as possible, and to keep the experiments computationally tractable.
%

Since even highly domain-specific corpora typically retain some lexical ambiguity (e.g., \emph{scale} in cooking still has readings related to measurement and fish skin), we cannot fully eliminate homonymy but aim to reduce it by restricting the domain.  
We settled on two corpora that met two additional criteria: in their unmodified state, the difference between homonyms and monosemes was clearly measurable by semantic neighborhood distance (cf.\ Tab.~\ref {tab:neighborhood}), and both use common, everyday language rather than specialized jargon, which supports the generalizability of our findings. The two corpora are (1) RecipeNLG \cite{bien-etal-2020-recipenlg}, containing cooking recipes (1.9 mio documents, 63k word types, 195 mio tokens), and (2) TinyStories \cite{eldan2023tinystories}, containing short LLM-generated children's stories in simple language (2.1 mio documents, 24k word types, 437 mio tokens). While RecipeNLG is more domain-specific, its vocabulary contains almost three times as many word types as Tiny Stories but only half as many tokens. For a language model, Tiny Stories, with its much lower type-token ratio, is easier to learn. 
\subsection{Modifying Corpora with Pseudowords}\label{sec:pseudo}
We modify each corpus with pseudowords that constitute either pseudo-homonyms or pseudo-hypernyms. Each corpus is modified with its own set of pseudowords, whose components are word pairs drawn from the respective corpus. The key variable underlying both types of pseudowords is the cosine distance between the components' embeddings: pseudo-homonyms are constructed from maximally dissimilar pairs, while pseudo-hypernyms are constructed from maximally similar ones (which are additionally filtered by taxonomic relatedness). This allows us to study whether the effects we observe depend on the semantic distance between the words that a pseudoword replaces. We proceed as follows: First, we compute the cosine distance for all word pairs in each corpus. We retain only pairs whose components each occur in at least 300 documents and whose frequencies differ by at most a factor of 10. We further require that each word appears as a component in at most one pseudoword, so that the resulting sets are non-overlapping.
For pseudo-\emph{homonyms}, we select the 50 most \emph{dissimilar} non-overlapping word pairs, representing words with semantically divergent readings (e.g., \emph{structure} and \emph{marmelade}).
For pseudo-\emph{hypernyms}, we select the most \emph{similar} non-overlapping word pairs and manually filter them for pairs that share a common category, i.e., pairs whose components could plausibly share a common ancestor in a taxonomy such as WordNet (e.g., \emph{mango} and \emph{papaya} are both fruit). This yields a list of 50 pseudo-hypernyms per corpus.
From each pair, we construct a pseudoword and manually create a name that is not part of the existing vocabulary, but remains reasonably similar to both components (e.g., \emph{marmelade} and \emph{structure} become \emph{structalade}). We verify that no pseudoword name occurs anywhere in the corpus, even as an incidental string. 

\subsection{Pseudowords in Semantic Space}\label{sec:space}
\begin{table}
\centering

\begin{tabular}{lcc}
\toprule
Ambiguity Type & \textsc{recipe} & \textsc{tiny} \\
\midrule
homonyms    & 0.28 & 0.32 \\
polysemes    & 0.24 & 0.30 \\
monosemes    & 0.21 & 0.28\\ \midrule
pseudo-homonyms & 0.36 & 0.40  \\
pseudo-hypernyms & 0.29 & 0.29 \\
\bottomrule
\end{tabular}
\caption{Mean neighborhood distance (20 nearest neighbors, Word2Vec) for standard homonyms, polysemes, and monosemes, and our pseudowords. A larger distance means that a word's neighbors are more spread out in semantic space.}
\label{tab:neighborhood}
\end{table}
\begin{figure*}
 \centering
 \includegraphics[scale=0.6]{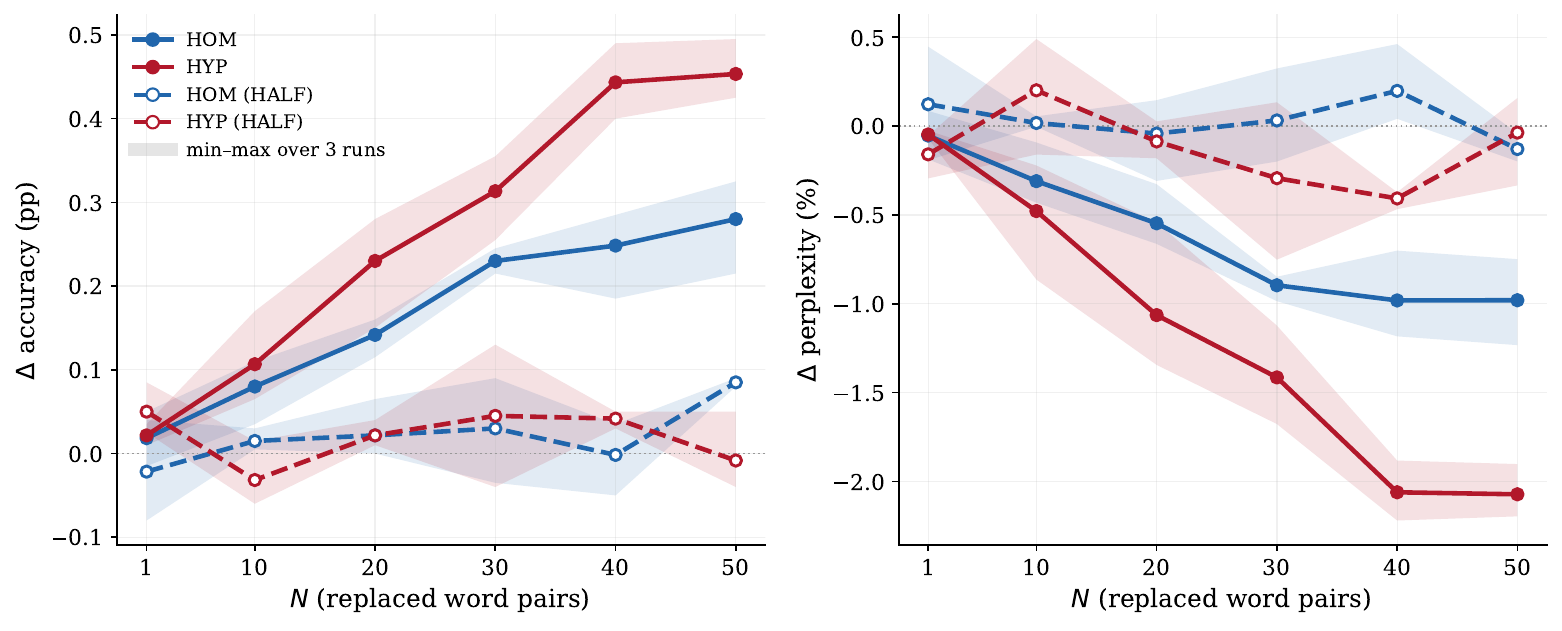}
 \caption{Changes in accuracy (left) and perplexity (right) compared to the unmodified base corpora after introducing N pseudo-homonyms (HOM) and pseudo-hypernyms (HYP), with 1--50 pseudoword types. Results are averaged over both corpora; regions indicate min/max over 3 runs. Changes for models with full component replacement are significant ($p<0.01$) from N=20, HALF models do not show significant changes.}\label{fig:recipecomp}
\end{figure*}
To verify that our pseudowords behave as expected, we compute the neighborhood distance for our pseudowords and compare them with standard monosemes, polysemes, and homonyms. For this, we use the list of homonyms, monosemes, and polysemes that \citet{https://doi.org/10.1111/cogs.12943} derived from the WordSmyth dictionary. To distinguish the three word types, Beekhuizen et al.\ measure the average pairwise cosine distance between a word and its nearest neighbors in a Word2Vec space \cite{NIPS2013_9aa42b31}. The core idea is that monosemes should have tight neighbor clusters with low average distance, while more ambiguous words will show higher average distances, because their readings have close neighbors in multiple regions of the semantic space. In their experiments, homonyms show the largest distances, which we can confirm in our corpora.
Table~\ref{tab:neighborhood} shows the average neighborhood distances using the 20 nearest neighbors for these word classes, and compares them with our pseudo-homonyms (HOM*) and pseudo-hypernyms (HYP*). As expected, our pseudo-homonyms show a neighborhood distance even larger than that of standard homonyms. The pseudo-hypernyms show a neighborhood distance comparable to that of natural homonyms in the recipe dataset: in Tiny Stories, it is closer to that of monosemes. This likely reflects the richer and more specialized vocabulary of the cooking domain, where co-hyponyms such as \emph{fridge} and \emph{oven} take on highly distinct functional roles and thus diverge more in semantic space than they would in a more general context.

\subsection{Corpus Variants for Ablations}\label{sec:variants}
We create multiple variants of our base datasets by replacing the component word pairs with pseudowords. We vary (1) the number of pseudoword types included (1, 10, 20, 30, 40, 50), (2) the type of pseudowords (pseudo-homonyms, "HOM" vs.\ pseudo-hypernyms, "HYP"), (3) whether we replace all occurrences of both components or only 50\% (HALF, with half the occurrences of each component being replaced), and (4) the base dataset. The HALF-condition creates a scenario closer to natural ambiguity and underspecification: in standard texts, many homonyms coexist with synonyms (e.g., \emph{bat} and \emph{club}), and hypernyms are used alongside their hyponyms. By including this condition, we can disentangle the effects of ambiguity and abstraction per se from the additional complexity introduced by the continued presence of the original component words. Overall, this yields 48 modified corpus variants and the two unmodified base corpora, for a total of 50 datasets, each serving as a training corpus for an individual language model. 
We show all pseudowords and their components along with word frequencies and cosine similarity in Appendix \ref{app:pseudowords}.

\section{Experiments and Evaluation}\label{sec:experiments}
\subsection{Models and Parameters}
For each of the 50 corpus variants (including the two unmodified base corpora), we train a GPT-2 model from scratch\footnote{\texttt{n\_layers=6, n\_head=6, n\_embed=384, dropout=0.2, learning\_rate=0.001, batch\_size=64, block\_size=256, grad\_clip=1.0}}. We emphasize that these are not fine-tuned from a pretrained checkpoint but trained from scratch on our corpora alone, so that any effect we observe can be attributed to the training data. For tokenization, we use the standard pre-trained GPT-2 BPE tokenizer across all models. Hyperparameters are kept identical across all models to avoid confounds. We train three runs per model configuration to account for variance. Using one NVIDIA A6000 GPU, each training run took between 25 and 40 minutes, for a total of approximately 75 GPU hours.
We evaluate each model by measuring next-token prediction accuracy on a held-out validation set. Each corpus variant has its own training, validation, and test split 
(80\%/15\%5\%), with the pseudoword modifications applied throughout. We first train and evaluate the baseline models on the unmodified corpora, then train individual models for each experimental condition and compare their next-token prediction accuracy. Significance is tested with a one-tailed t-test, averaging over the three runs per configuration; when aggregating conditions, we average over all runs within the group.

\subsection{Increasing Ambiguity and Underspecification}\label{sec:performance}
In our first set of experiments, we successively introduce pseudo-homonyms and pseudo-hypernyms into each dataset, replacing all occurrences of the component words. We increase the number of pseudoword types starting with 1, then from 10 to 50 in increments of 10. This simulates increased ambiguity and underspecification in the models' language and allows us to compare the two. Apart from the first word (where we use the most similar component word pair for HYP and the most dissimilar one for HOM), the average cosine distance of the pseudoword components is constant across all models.
%

Figure \ref{fig:recipecomp} shows the results for all runs, averaged over both corpora (cf.~App.~\ref{app:perf} for individual corpus results).
The introduction of greater ambiguity and underspecification improves performance if the new words replace both components (i.e., there is only "papango" left in the corpus, and both "mango" and "papaya" are replaced in all occurrences). If we retain the components (equivalent to synonyms for the pseudo-homonyms and hyponyms for the pseudo-hypernyms), there are no significant performance changes.  

\begin{figure}
 \centering
 \includegraphics[scale=0.5]{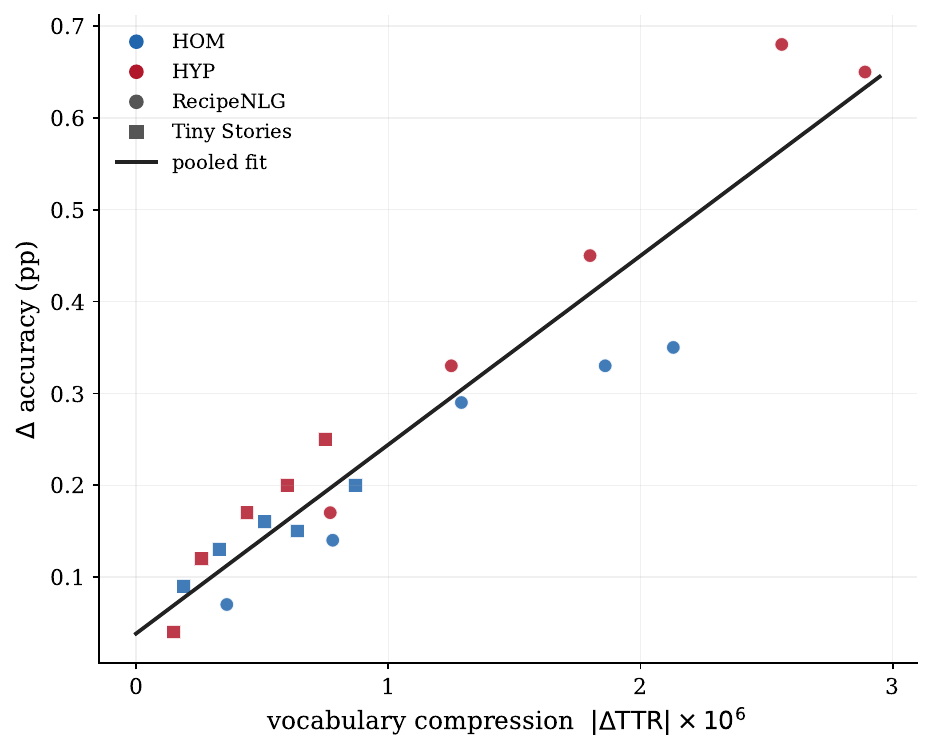}
 \caption{Linear regression predicting changes in LM accuracy from TTR for all models; regression coefficient for TTR change is 0.86.}\label{fig:ttr_predict}
\end{figure}

Changing the level of ambiguity or underspecification also changes the type-token ratio (TTR) of the underlying corpus (more pseudoword types $\rightarrow$ smaller TTR). According to the optimality theory of language, this vocabulary compression licenses ambiguity as a means of efficiency.  To test whether this also predicts model performance, we compute TTR based on the corpus tokenized with the BPE-tokenizer used to train the models, defined as the ratio of unique types to the total number of tokens. We then train a linear regression model to predict the increase in LM accuracy from the reduction in TTR. Fig.~\ref{fig:ttr_predict} shows the results.

We find that we can almost perfectly predict the change in accuracy from the change in TTR, which also explains the differences between HOM and HYP. Across all models and conditions, the regression coefficient is 0.86; excluding the HALF models (which show no significant change in accuracy), the coefficient is even higher at 0.96. 

Note that the change in TTR is driven not only by the reduction in token types but also by an increase in the absolute number of tokens (cf.~App.\ref{app:perf}), because the pseudowords are all tokenized by the BPE tokenizer. Our setup thus does not gain resource efficiency for generation, just a more compact vocabulary. Exploring whether using in-vocabulary words rather than pseudowords could provide a valid mechanism for model compression is a subject for future work.

\emph{Takeaways: We see linguistic optimality theory reflected in overall model performance: By compressing the language using homonyms and hypernyms, models become more effective. Neither ambiguity nor abstraction adds complexity that outweighs the gains from a smaller type-token ratio.}

\subsection{Pseudowords and Their Components}\label{sec:components}
\begin{figure}
\centering
\includegraphics[scale=0.5]{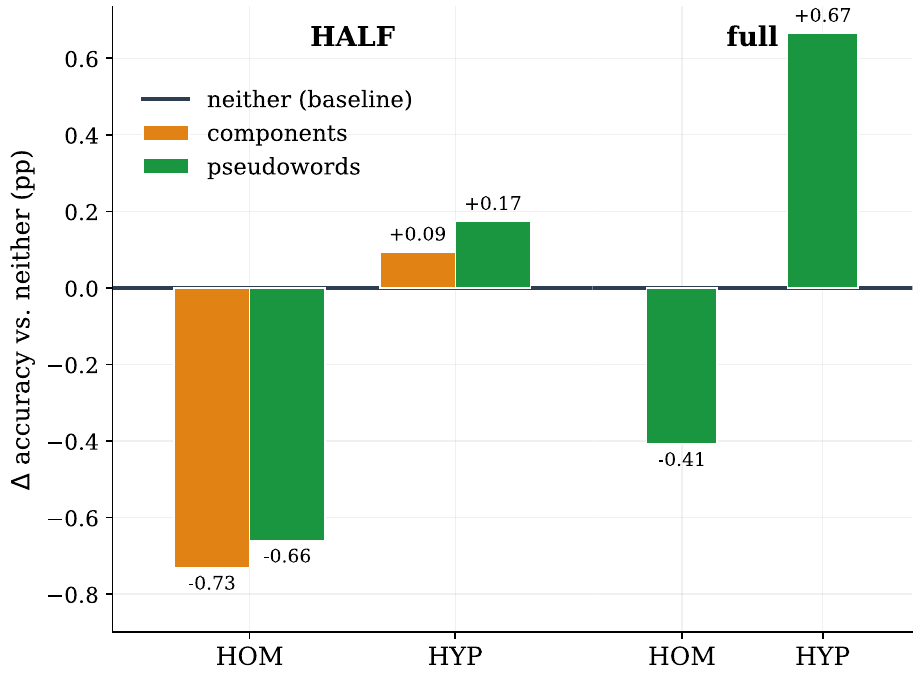}
\caption{Comparing accuracy for batches containing pseudowords or components with batches containing neither of them, averaged over both corpora and either all HALF-models (left) or models with full replacement (right), measured as delta with other batches ("neither"). All results, except HYP in HALF models, are significant ($p<0.01$).}\label{fig:pseudocomp}
\end{figure}

While the previous experiment showed that the presence of synonyms and hyponyms (in the HALF models) makes the language harder for the model to learn, overall performance did not decrease. We want to explore the results in more detail, assessing how well the pseudowords and their synonyms (or hyponyms) can be generated. For this, we separate the validation batches into three groups: (1) batches containing at least one pseudoword, (2) batches containing at least one pseudoword component but no pseudoword (for the HALF models only), and (3) batches containing neither. This lets us assess whether the model can reliably generate pseudowords, whether accuracy on surrounding tokens is affected, and whether component words become harder to generate when competing with a pseudo-homonym or pseudo-hypernym. Note that we do not directly compare the accuracy of producing pseudowords or their components, since single-word accuracy is difficult to match with adequate controls. Instead, we use the accuracy on batches containing the target words as a proxy for the model's ability to generate them.
Figure~\ref{fig:pseudocomp} shows the results. We report the difference in accuracy between pseudoword batches or component batches (HALF models only) with batches containing no pseudowords or components ("neither"). Batches containing pseudo-homonyms or their components show significantly lower accuracy than batches without them. Pseudo-hypernyms, by contrast, perform \emph{better} on both pseudoword batches and component batches, with significant increases for pseudo-hypernyms in the full replacement strategy. In all cases, the presence of synonyms or hyponyms of the pseudowords lowers performance, with the component words performing even worse than the pseudowords themselves. This mechanism could, at a larger scale, contribute to bias — for instance, introducing a general term like \emph{president} could make more specific terms like \emph{chairman} or \emph{chairwoman} harder to learn.

\emph{Takeaways: Pseudo-homonyms are hard to learn, but pseudo-hypernyms can be generated reliably. In all cases, the presence of component words lowers performance. While ambiguity is not challenging per se, its integration in language with synonyms adds complexity.}
\subsection{Findings on localization}\label{sec:localize}
In addition to measuring model performance, we analyzed model parameters to provide a mechanistic account of ambiguity and underspecification. We examine whether any geometric measures vary linearly with the number of pseudowords. We mainly focused on Von Neumann Entropy (VNE), which is the Shannon Entropy of a matrix's eigenvalues. We hypothesized that greater ambiguity or underspecification would increase entropy, showing higher uncertainty in the model. We indeed found that the effective rank of the first weight matrix (usually denoted $W_1$) in the Multilayer Perceptron (MLP) increases linearly with the number of pseudowords (details in App.~\ref{app:vne}). This increase is independent of type-token ratio: It happens in the HALF-models and in all HOM and HYP models. Recent studies have identified MLP layers as effective predictors of diverse semantic relations \cite{icml26_svd}.  However, we find significant effects only for RecipeNLG, not for Tiny Stories (where changes in TTR and performance were also much smaller than for RecipeNLG).  While we consider this a preliminary finding, the localization in the mid-layers aligns well with previous work and our own insights into disambiguation (cf. Sec.~\ref{sec:dis}): Most of the changes in VNE were observable in the middle layers, where we also observe disambiguation.
We found no difference in the input embeddings, neither in VNE nor in the spread of the vector space used. While other measures might reveal more fine-grained differences here, we assume that the parameters change only where actual disambiguation occurs. 

\subsection{Disambiguation and Specification}\label{sec:dis}
With probing experiments, we want to learn whether the internal processing of ambiguity differs from that of underspecification, and whether we can locate the disambiguation and sense specification within the model activations. 
We probe the four *50-HALF models using the pseudowords and their component words for disambiguation or specification. For this, we use lexical patterns with either ambiguous or disambiguated sentences. We measure whether the model-internal activations change when we add the component word that disambiguates the sense (a synonym) or specifies it (a hyponym). E.g., a neutral pattern is \texttt{This is another [structalade], like the other one.} (inserting the pseudoword), and the corresponding disambiguated pattern would be \texttt{This is another [structalade], like the other [marmelade].} (inserting both pseudoword and one of the component words). As a reference for component-word activation, we use the neutral pattern and insert the component word in place of the pseudoword (e.g., \texttt{This is another [marmelade], like the other one}). 

\begin{figure}
\centering
\includegraphics[scale=0.45]{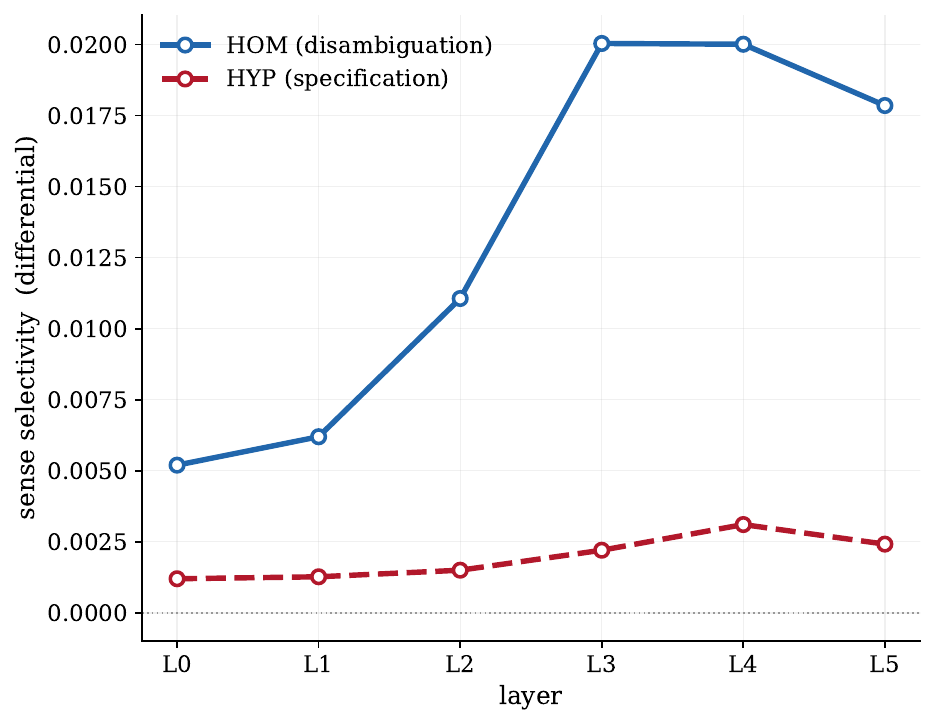}
\caption{Sense selectivity in the 50-HALF models. Pseudo-homonyms are disambiguated; pseudo-hypernyms remain underspecified.}\label{fig:disambiguation}

\end{figure}

We use 5 different pattern pairs (cf. App.~\ref{app:dis}) and compute the internal activations for the sequences at the sentence-end position after each layer. We then measure how much the activation of the neutral pseudoword pattern shifts toward the neutral component-word patterns upon disambiguation by calculating the differential diff as follows, with $x$ denoting pseudowords, $y$ denoting components, and $_n$ denoting the neutral (0) pattern or the pattern disambiguated with components 1 or 2.
\begin{multline}
\footnotesize
\mathrm{diff} = \frac{1}{2}[
  (\cos(\mathbf{x}_{y_1}, \mathbf{y}_1) - \cos(\mathbf{x}_0, \mathbf{y}_1))  \\
\footnotesize + (\cos(\mathbf{x}_{y_2}, \mathbf{y}_2) - \cos(\mathbf{x}_0, \mathbf{y}_2))
\footnotesize]\end{multline}
The results are shown in Figure~\ref{fig:disambiguation}. The activations for the pseudo-hypernyms do not change at all in the context of one of their hyponyms. The pseudo-homonyms, in contrast, are disambiguated in the middle layers, decreasing the distance to the most similar sense of their synonym component. More detailed inspections also reveal that the pseudo-homonyms, as expected, show a sense bias toward the more frequent component, measured as the cosine similarity between the neutral patterns of the pseudowords and the same pattern instantiated with the component word (cf. App.~\ref{app:dis}). 
This means that the pseudo-hypernyms hardly ever shift and remain underspecified. This could be either because the component words are very similar to the pseudo-hypernym (making the difference hard to detect) or because a model-internal mechanism retains underspecification when further specification does not influence follow-up generation. While we cannot conclude that hypernyms are never specified in the model (we did see that hyponyms can be reliably generated, cf. Sec.~\ref{sec:components}), it might mean that for the model, disambiguation is always necessary, while underspecification can be retained. Future work addressing bias arising from abstraction errors could explore and manipulate this mechanism to avoid potentially harmful underspecification.

\emph{Takeaways: Disambiguation of homonyms is reflected in the model activations (especially in the middle layers), while hypernyms remain underspecified.}
\section{Conclusion} \label{sec:conclusion}
Our study has shown how increasing ambiguity or underspecification changes the behavior of language models trained on this data, and how these two phenomena differ. We evaluated changes in LM performance by using pseudo-homonyms or pseudo-hypernyms. We showed that increasing either ambiguity or underspecification improves overall model performance, a pattern that can be directly predicted from the decrease in the type-token ratio. In that respect, ambiguity and underspecification have similar effects in language models compared to those predicted for human language processing. However, we find that in settings where synonyms or individual hyponyms still occur in the training corpus, both the pseudowords and their component words are generated less reliably than other vocabulary. We also provided a preliminary mechanistic account of ambiguity, underspecification, and disambiguation: more ambiguous or underspecified language increases Von-Neumann Entropy in the model's MLP layer parameters. Further, activation-based disambiguation analysis shows that the readings of pseudo-homonyms are resolved, but underspecification for pseudo-hypernyms is retained.

We observed significant effects with concrete implications for responsible LLM design. On the one hand, model vocabulary can be compressed by reducing vocabulary diversity through abstraction, leading to higher overall performance. On the other hand, ambiguity can harm performance for both the underspecified expressions and their hyponyms or individual readings. 

In future work, we aim to explore the effects of ambiguity on a larger scale, and provide more measures for localizing ambiguity within the model. Further, we aim to dig deeper into the processes of resolving ambiguity and underspecification, giving a more precise account of when a resolution happens and when it does not. Those studies can then help to anchor bias research in mechanistic interpretability for fairer and more reliable LLMs. 
\section*{Limitations}
Our experiments comprise many individual language models based on several modifications of two base corpora. While this provides an in-depth analysis, the scope is limited in several respects:
\paragraph{Model choice:} All our models are GPT-2 models trained from scratch. While we believe that our results carry over to other model families, this remains to be verified. Further, to avoid confounders, we did not optimize hyperparameters individually for each run; some effects might change with hyperparameter tuning.
\paragraph{Ambiguity and abstraction:} With pseudohomonyms and pseudohypernyms, we study only one type of ambiguity and one type of abstraction. Both phenomena occur at multiple linguistic levels that interact with each other, and we have only examined a small subset that we could cover with pseudowords. Future work will need to present experimental setups that accommodate different types of ambiguity, such as syntactic or discourse-level ambiguity.
\paragraph{Scaling and evaluation:} To isolate effects of small language changes, we used small language models and restricted corpora. Future work needs to scale our findings to larger models, different domain-specific corpora, and more general corpora. In those settings, established benchmarks (e.g., HellaSwag) can be used to test the impact of ambiguity; such benchmarks are not meaningful for our small models, so we rely on next-token prediction accuracy as our sole evaluation metric. Scaling up will also show whether our assumption holds that models compressed through pseudohypernyms still perform well on downstream tasks.
\section*{Ethical Considerations}
Our research can contribute to more sustainable model development: we provide initial evidence that linguistically motivated compression of LLMs is feasible, showing that pseudohypernyms can reduce vocabulary variety and improve performance, especially for domain-specific models. If these findings scale, they could reduce the computational resources and energy required for model training. We also believe our results can inform research on fairness, since we show that the introduction of ambiguous or underspecified terms can degrade accuracy not only for those terms but also for their related words — a mechanism that, at scale, could contribute to bias propagation in ways that are difficult to detect by examining individual terms in isolation.

\bibliography{ambiguity}
\bibliographystyle{acl_natbib}

\clearpage
\appendix
\onecolumn
\section{Pseudowords}\label{app:pseudowords}
\begin{table}[H]
\centering
\small
\begin{tabular}{llrlrlr}
\toprule
Pseudoword & Word 1 & Freq & Word 2 & Freq & Cos.\ Sim. & Total Freq \\
\midrule
simmoil & boil & 465,659 & simmer & 365,022 & 0.729 & 830,681 \\
liemon & lemon & 268,966 & lime & 80,588 & 0.781 & 349,554 \\
poreef & beef & 191,114 & pork & 146,549 & 0.752 & 337,663 \\
dryet & dry & 244,075 & wet & 34,358 & 0.737 & 278,433 \\
poettle & pot & 255,925 & kettle & 11,061 & 0.823 & 266,986 \\
thinick & thick & 144,560 & thin & 77,768 & 0.728 & 222,328 \\
plaer & plate & 140,468 & platter & 56,274 & 0.751 & 196,742 \\
freezidge & freezer & 48,748 & fridge & 37,587 & 0.771 & 86,335 \\
glaramic & glass & 78,152 & ceramic & 3,763 & 0.715 & 81,915 \\
muffcake & muffin & 67,858 & cupcake & 13,692 & 0.822 & 81,550 \\
salmout & salmon & 58,542 & trout & 4,306 & 0.835 & 62,848 \\
caramudge & caramel & 39,598 & fudge & 22,149 & 0.711 & 61,747 \\
refrense & rinse & 59,056 & refresh & 1,670 & 0.710 & 60,726 \\
jattle & jar & 37,238 & bottle & 9,179 & 0.738 & 46,417 \\
grerry & cherry & 34,893 & grape & 6,964 & 0.719 & 41,857 \\
blulse & pulse & 37,911 & blitz & 1,476 & 0.807 & 39,387 \\
peaspberry & peach & 20,202 & raspberry & 16,284 & 0.745 & 36,486 \\
oatsin & oatmeal & 25,461 & raisin & 8,368 & 0.761 & 33,829 \\
jarmelade & jam & 25,980 & marmalade & 6,673 & 0.746 & 32,653 \\
bluehubarb & blueberry & 16,490 & rhubarb & 15,525 & 0.725 & 32,015 \\
panaffle & pancake & 14,892 & waffle & 7,896 & 0.730 & 22,788 \\
yellurple & yellow & 18,831 & purple & 1,908 & 0.721 & 20,739 \\
healsty & healthy & 12,743 & tasty & 7,476 & 0.817 & 20,219 \\
papango & mango & 17,207 & papaya & 2,854 & 0.904 & 20,061 \\
lieal & veal & 13,098 & liver & 5,539 & 0.736 & 18,637 \\
textearrance & texture & 16,686 & appearance & 1,337 & 0.740 & 18,023 \\
sodiprot & protein & 10,322 & sodium & 7,533 & 0.777 & 17,855 \\
breakfrunch & breakfast & 14,727 & brunch & 2,537 & 0.855 & 17,264 \\
padook & paddle & 10,457 & hook & 5,257 & 0.785 & 15,714 \\
champodka & vodka & 8,281 & champagne & 4,945 & 0.715 & 13,226 \\
caviteck & cavity & 8,407 & neck & 4,415 & 0.726 & 12,822 \\
summinter & summer & 9,199 & winter & 3,509 & 0.833 & 12,708 \\
outinner & outer & 9,098 & inner & 2,565 & 0.757 & 11,663 \\
pepperami & pepperoni & 7,960 & salami & 3,482 & 0.832 & 11,442 \\
elacky & elastic & 9,252 & tacky & 1,157 & 0.757 & 10,409 \\
braerm & bran & 7,327 & germ & 2,945 & 0.726 & 10,272 \\
octoquid & squid & 4,773 & octopus & 1,884 & 0.827 & 6,657 \\
sanirt & sand & 3,084 & dirt & 3,023 & 0.727 & 6,107 \\
stalulb & stalk & 2,623 & bulb & 2,602 & 0.730 & 5,225 \\
sklank & flank & 4,018 & skirt & 1,090 & 0.888 & 5,108 \\
elemource & source & 3,850 & element & 1,070 & 0.777 & 4,920 \\
vitcium & vitamin & 2,349 & calcium & 1,608 & 0.890 & 3,957 \\
outinoor & outdoor & 2,974 & indoor & 787 & 0.834 & 3,761 \\
parazontal & parallel & 1,214 & horizontal & 1,117 & 0.787 & 2,331 \\
cartoub & tub & 1,174 & carton & 1,113 & 0.771 & 2,287 \\
intextrior & interior & 1,383 & exterior & 893 & 0.810 & 2,276 \\
shrillapse & shrink & 1,376 & collapse & 631 & 0.739 & 2,007 \\
leapill & spill & 825 & leak & 714 & 0.718 & 1,539 \\
cleammer & cleaver & 736 & hammer & 714 & 0.761 & 1,450 \\
cuptry & pantry & 791 & cupboard & 372 & 0.792 & 1,163 \\
\bottomrule
\end{tabular}\caption{Pseudoword mappings for hypernyms on RecipeNLG, including word frequencies on the raw corpus and cosine similarity of the component words.}\label{tab:hypernyms-recipe}
\end{table}

\begin{table*}[t]
\centering
\small

\begin{tabular}{llrlrlr}
\toprule
Pseudoword & Word 1 & Freq & Word 2 & Freq & Cos.\ Sim. & Total Freq \\
\midrule
wilookie & cookie & 153,461 & wild & 9,416 & -0.170 & 162,877 \\
livide & divide & 83,732 & lid & 60,620 & -0.147 & 144,352 \\
wintress & press & 117,501 & winter & 3,509 & -0.163 & 121,010 \\
strorrange & arrange & 106,223 & strong & 3,642 & -0.163 & 109,865 \\
microwight & microwave & 103,087 & slight & 3,413 & -0.158 & 106,500 \\
glaender & glass & 78,152 & render & 1,902 & -0.145 & 80,054 \\
wooxact & wooden & 44,176 & exact & 1,168 & -0.140 & 45,344 \\
speace & spice & 32,787 & spear & 1,447 & -0.150 & 34,234 \\
bittachment & bite & 18,767 & attachment & 15,049 & -0.159 & 33,816 \\
dontire & dozen & 18,542 & entire & 15,252 & -0.160 & 33,794 \\
cheeseboo & cheesecake & 27,700 & bamboo & 4,777 & -0.141 & 32,477 \\
goist & moist & 20,544 & gas & 11,511 & -0.141 & 32,055 \\
critamin & crab & 26,918 & vitamin & 2,349 & -0.146 & 29,267 \\
marrelly & jelly & 25,852 & marrow & 708 & -0.160 & 26,560 \\
quoal & quart & 25,070 & goal & 712 & -0.144 & 25,782 \\
abutty & above & 20,012 & nutty & 3,633 & -0.140 & 23,645 \\
cruink & crush & 19,653 & link & 1,200 & -0.147 & 20,853 \\
looduct & loosen & 17,763 & product & 2,650 & -0.163 & 20,413 \\
goassel & goat & 14,969 & vessel & 1,148 & -0.144 & 16,117 \\
rollider & cider & 14,321 & roller & 747 & -0.204 & 15,068 \\
cartine & pine & 13,844 & carton & 1,113 & -0.146 & 14,957 \\
ripoodle & noodle & 11,801 & ripe & 2,826 & -0.140 & 14,627 \\
warmquake & warmed & 13,740 & earthquake & 575 & -0.208 & 14,315 \\
pufoody & puff & 13,034 & bloody & 1,154 & -0.154 & 14,188 \\
clundae & cloth & 11,798 & sundae & 1,357 & -0.148 & 13,155 \\
jeroam & foam & 9,849 & jerk & 1,816 & -0.162 & 11,665 \\
distribaft & distribute & 10,756 & draft & 769 & -0.156 & 11,525 \\
ranome & ranch & 9,811 & dome & 842 & -0.178 & 10,653 \\
drulastic & elastic & 9,252 & drunken & 570 & -0.151 & 9,822 \\
reackie & reach & 8,389 & quickie & 339 & -0.146 & 8,728 \\
threapinion & thread & 8,363 & opinion & 320 & -0.158 & 8,683 \\
scramender & scramble & 4,586 & lavender & 2,898 & -0.149 & 7,484 \\
baskush & basket & 6,754 & lush & 324 & -0.158 & 7,078 \\
structalade & marmalade & 6,673 & structure & 334 & -0.162 & 7,007 \\
originield & original & 5,793 & shield & 411 & -0.174 & 6,204 \\
nimell & smell & 4,381 & nine & 1,675 & -0.181 & 6,056 \\
aceadball & meatball & 4,434 & acid & 1,226 & -0.140 & 5,660 \\
purpograph & photograph & 2,928 & purpose & 2,033 & -0.147 & 4,961 \\
boultry & poultry & 4,602 & bounce & 315 & -0.146 & 4,917 \\
clioneer & cling & 3,650 & pioneer & 358 & -0.155 & 4,008 \\
shelitation & shelf & 3,184 & imitation & 703 & -0.165 & 3,887 \\
washtball & washed & 2,513 & football & 656 & -0.147 & 3,169 \\
comfold & unfold & 2,269 & comfort & 725 & -0.143 & 2,994 \\
taucy & saucy & 2,161 & tape & 685 & -0.187 & 2,846 \\
elerune & prune & 1,680 & element & 1,070 & -0.147 & 2,750 \\
bimpfire & bind & 1,125 & campfire & 814 & -0.145 & 1,939 \\
stourve & stout & 978 & curve & 874 & -0.171 & 1,852 \\
shortcloss & gloss & 1,007 & shortcut & 560 & -0.169 & 1,567 \\
miccasional & minus & 811 & occasional & 311 & -0.139 & 1,122 \\
filliss & file & 785 & bliss & 334 & -0.144 & 1,119 \\
\bottomrule
\end{tabular}\caption{Pseudoword mappings for homonyms on RecipeNLG, including word frequencies on the raw corpus and cosine similarity of the component words.}
\label{tab:homonyms-recipe}
\end{table*}

\begin{table*}[t]
\centering
\small

\begin{tabular}{llrlrlr}
\toprule
Pseudoword & Word 1 & Freq & Word 2 & Freq & Cos.\ Sim. & Total Freq \\
\midrule
fimp & fish & 331,746 & shrimp & 16,052 & 0.878 & 347,798 \\
wirty & wet & 111,774 & dirty & 92,013 & 0.826 & 203,787 \\
harf & hat & 159,558 & scarf & 28,519 & 0.837 & 188,077 \\
purplink & pink & 87,515 & purple & 57,877 & 0.894 & 145,392 \\
towack & tower & 76,352 & stack & 12,493 & 0.831 & 88,845 \\
fanch & farm & 69,515 & ranch & 9,316 & 0.927 & 78,831 \\
rivetream & river & 60,871 & stream & 14,907 & 0.906 & 75,778 \\
princeen & princess & 42,117 & queen & 31,142 & 0.840 & 73,259 \\
spoork & spoon & 39,697 & fork & 16,591 & 0.854 & 56,288 \\
shoat & sheep & 34,461 & goat & 21,558 & 0.839 & 56,019 \\
chickig & pig & 33,503 & chicken & 21,253 & 0.836 & 54,756 \\
comeoon & moon & 37,477 & comet & 16,536 & 0.828 & 54,013 \\
smilthy & smelly & 33,591 & filthy & 19,034 & 0.823 & 52,625 \\
druitar & drum & 30,649 & guitar & 20,696 & 0.902 & 51,345 \\
glape & tape & 25,934 & glue & 25,356 & 0.887 & 51,290 \\
cirquare & circle & 27,623 & square & 20,474 & 0.878 & 48,097 \\
expenseap & expensive & 26,239 & cheap & 20,848 & 0.824 & 47,087 \\
mitir & mix & 32,978 & stir & 13,886 & 0.832 & 46,864 \\
sugutter & sugar & 25,977 & butter & 20,459 & 0.845 & 46,436 \\
genetful & generous & 25,709 & thoughtful & 18,458 & 0.865 & 44,167 \\
bromb & brush & 25,528 & comb & 17,488 & 0.821 & 43,016 \\
liemon & lemon & 26,872 & lime & 14,473 & 0.897 & 41,345 \\
caramp & card & 25,641 & stamp & 13,903 & 0.852 & 39,544 \\
soampoo & soap & 25,445 & shampoo & 11,783 & 0.843 & 37,228 \\
sharctopus & shark & 23,177 & octopus & 13,818 & 0.856 & 36,995 \\
swield & sword & 21,310 & shield & 13,959 & 0.876 & 35,269 \\
jumu & jug & 17,897 & mug & 16,021 & 0.818 & 33,918 \\
violute & flute & 18,008 & violin & 15,442 & 0.952 & 33,450 \\
pencebook & pencil & 18,854 & notebook & 14,368 & 0.836 & 33,222 \\
potanion & potato & 16,314 & onion & 16,203 & 0.898 & 32,517 \\
bable & barn & 22,537 & stable & 9,804 & 0.838 & 32,341 \\
crocogator & crocodile & 17,066 & alligator & 14,489 & 0.943 & 31,555 \\
threedle & thread & 15,857 & needle & 15,485 & 0.848 & 31,342 \\
hammail & hammer & 18,172 & nail & 12,948 & 0.813 & 31,120 \\
scootorcycle & scooter & 15,412 & motorcycle & 14,913 & 0.823 & 30,325 \\
collash & collar & 17,038 & leash & 12,855 & 0.831 & 29,893 \\
chimnoof & roof & 16,375 & chimney & 11,770 & 0.826 & 28,145 \\
waffin & muffin & 14,239 & waffle & 13,755 & 0.888 & 27,994 \\
deackboard & desk & 15,800 & blackboard & 12,124 & 0.811 & 27,924 \\
borrend & borrow & 16,888 & lend & 10,390 & 0.816 & 27,278 \\
fooseball & football & 13,762 & baseball & 12,400 & 0.942 & 26,162 \\
thundning & thunder & 13,807 & lightning & 11,450 & 0.869 & 25,257 \\
peacrune & peach & 15,844 & prune & 8,209 & 0.888 & 24,053 \\
skouse & skirt & 11,618 & blouse & 10,851 & 0.958 & 22,469 \\
hubin & hut & 11,438 & cabin & 10,424 & 0.910 & 21,862 \\
pangaroo & kangaroo & 11,222 & panda & 10,330 & 0.840 & 21,552 \\
sunriet & sunset & 9,485 & sunrise & 8,504 & 0.868 & 17,989 \\
scoal & goal & 10,024 & score & 7,627 & 0.834 & 17,651 \\
juicipe & juicy & 12,090 & ripe & 2,029 & 0.814 & 14,119 \\
slimble & slip & 7,898 & stumble & 2,398 & 0.830 & 10,296 \\
\bottomrule
\end{tabular}\caption{Pseudoword mappings for hypernyms on TinyStories, including word frequencies on the raw corpus and cosine similarity of the component words.}
\label{tab:hypernyms-tiny}
\end{table*}

\begin{table*}[h]
\centering
\small

\begin{tabular}{llrlrlr}
\toprule
Pseudoword & Word 1 & Freq & Word 2 & Freq & Cos.\ Sim. & Total Freq \\
\midrule
slark & dark & 118,492 & slice & 3,691 & -0.162 & 122,183 \\
earthquink & pink & 87,515 & earthquake & 6,802 & -0.188 & 94,317 \\
wrise & wise & 89,254 & wrist & 2,511 & -0.172 & 91,765 \\
colorision & colorful & 85,866 & decision & 3,184 & -0.193 & 89,050 \\
harair & chair & 78,373 & harm & 1,796 & -0.169 & 80,169 \\
bitinue & bite & 60,784 & continue & 3,010 & -0.192 & 63,794 \\
singush & push & 49,699 & singer & 2,066 & -0.209 & 51,765 \\
pashlight & pig & 33,503 & flashlight & 16,743 & -0.178 & 50,246 \\
avorthday & birthday & 46,154 & avoid & 1,571 & -0.218 & 47,725 \\
brealthy & breath & 21,775 & wealthy & 21,283 & -0.170 & 43,058 \\
alermon & lemon & 26,872 & alert & 15,862 & -0.162 & 42,734 \\
spraleep & spray & 20,532 & asleep & 20,288 & -0.204 & 40,820 \\
flitten & kitten & 39,322 & flip & 1,331 & -0.166 & 40,653 \\
flammy & flag & 25,002 & tummy & 14,339 & -0.236 & 39,341 \\
yierror & mirror & 26,403 & yield & 8,162 & -0.212 & 34,565 \\
drice & dig & 26,601 & price & 6,436 & -0.167 & 33,037 \\
finettle & finger & 21,459 & settle & 7,913 & -0.166 & 29,372 \\
deighbour & dive & 14,339 & neighbor & 12,812 & -0.189 & 27,151 \\
twinkose & hose & 22,966 & twinkle & 3,903 & -0.170 & 26,869 \\
massareless & careless & 19,120 & massage & 7,237 & -0.184 & 26,357 \\
flaesure & measure & 22,438 & flash & 3,090 & -0.167 & 25,528 \\
decoulder & decide & 15,754 & shoulder & 8,657 & -0.185 & 24,411 \\
creamare & scare & 22,658 & creamy & 1,334 & -0.173 & 23,992 \\
stearrot & parrot & 18,697 & steady & 2,323 & -0.207 & 21,020 \\
repiggle & repair & 12,499 & giggle & 8,461 & -0.172 & 20,960 \\
stickut & hut & 11,438 & sticker & 8,885 & -0.182 & 20,323 \\
goldath & math & 14,178 & golden & 5,470 & -0.224 & 19,648 \\
spaghettadge & spaghetti & 16,822 & badge & 2,143 & -0.173 & 18,965 \\
remandbox & sandbox & 12,385 & remind & 6,274 & -0.184 & 18,659 \\
elecious & elevator & 14,637 & precious & 3,505 & -0.198 & 18,142 \\
recipull & bull & 16,384 & recipe & 1,472 & -0.222 & 17,856 \\
screasure & screen & 15,820 & pleasure & 1,400 & -0.178 & 17,220 \\
wintratch & winter & 10,685 & scratch & 6,296 & -0.174 & 16,981 \\
twust & dust & 12,846 & twin & 4,060 & -0.169 & 16,906 \\
dioldier & soldier & 15,385 & dip & 1,392 & -0.173 & 16,777 \\
clorehead & club & 14,396 & forehead & 1,807 & -0.176 & 16,203 \\
firepleeper & fireplace & 11,799 & keeper & 3,516 & -0.174 & 15,315 \\
helpace & helper & 11,184 & surface & 2,154 & -0.183 & 13,338 \\
overcray & ashtray & 11,847 & overcome & 1,382 & -0.243 & 13,229 \\
spail & rail & 9,588 & spit & 3,464 & -0.168 & 13,052 \\
subful & subway & 11,163 & handful & 1,652 & -0.167 & 12,815 \\
invick & invite & 10,202 & click & 1,830 & -0.166 & 12,032 \\
steader & sticky & 8,169 & leader & 3,808 & -0.236 & 11,977 \\
distrand & disturb & 9,237 & brand & 2,432 & -0.197 & 11,669 \\
pativer & shiver & 8,476 & patience & 3,160 & -0.163 & 11,636 \\
deseak & leak & 8,937 & deserve & 2,006 & -0.194 & 10,943 \\
applackpack & backpack & 5,766 & applaud & 4,967 & -0.191 & 10,733 \\
embooden & embarrass & 4,988 & wooden & 3,956 & -0.176 & 8,944 \\
luxureel & luxury & 5,381 & peel & 2,116 & -0.205 & 7,497 \\
snowbey & snowball & 3,976 & obey & 2,763 & -0.176 & 6,739 \\
\bottomrule
\end{tabular} \caption{Pseudoword mappings for homonyms on TinyStories, including word frequencies on the raw corpus and cosine similarity of the component words}
\label{tab:homonyms-tiny}
\end{table*}

\onecolumn
\section{Model performance split by corpora}\label{app:perf}
\begin{figure}[H]
 \centering
 \includegraphics[scale=0.5]{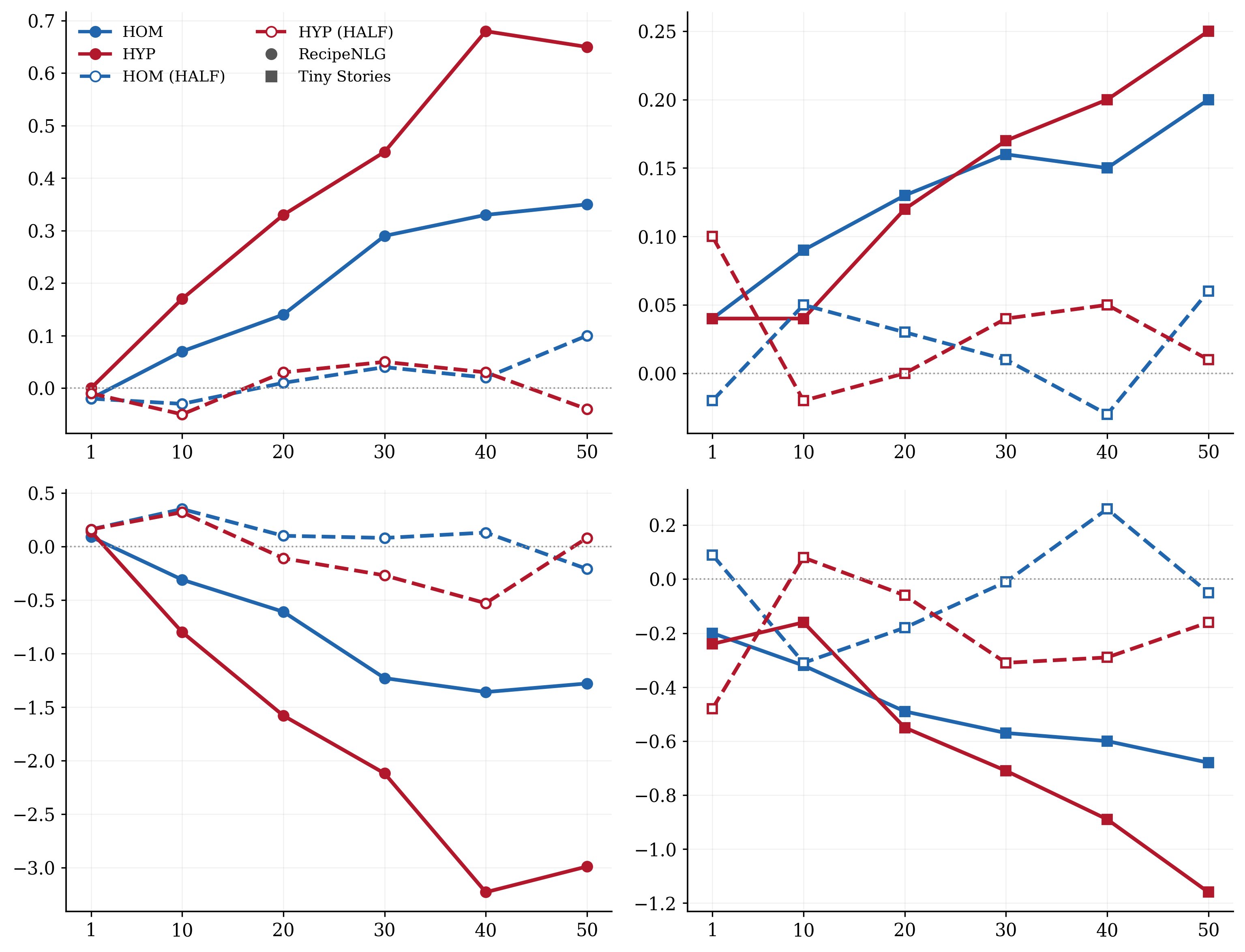}
 \caption{Accuracy (upper) and Perplexity (lower) changes in the models for increasing ambiguous (HOM) or underspecified (HYP) pseudowords, split by corpus. }
\vspace*{5mm}
 \includegraphics[scale=0.3]{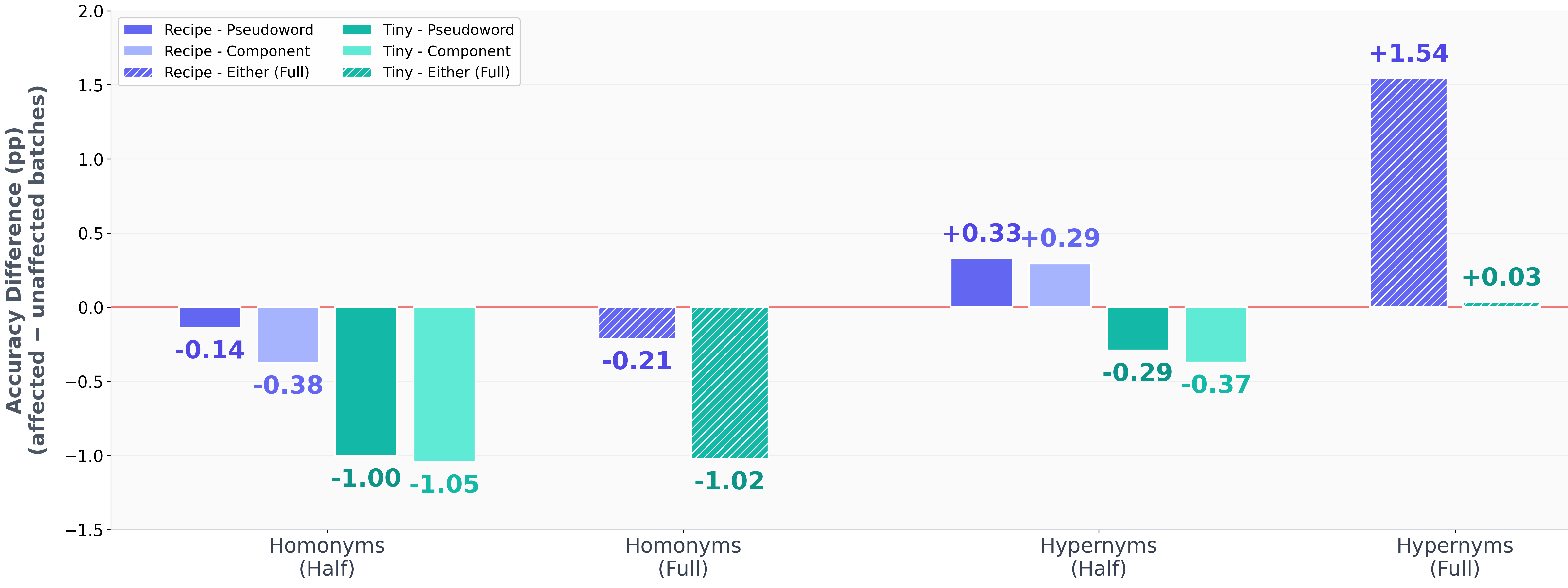}
 \caption{Model performance for generating batches with either pseudowords or component words, split by corpora. Bars indicate differences to batches that contain neither pseudowords nor components.}\label{fig:comp_split}
\end{figure}
\begin{figure}[H]
\centering
 \includegraphics[scale=0.5]{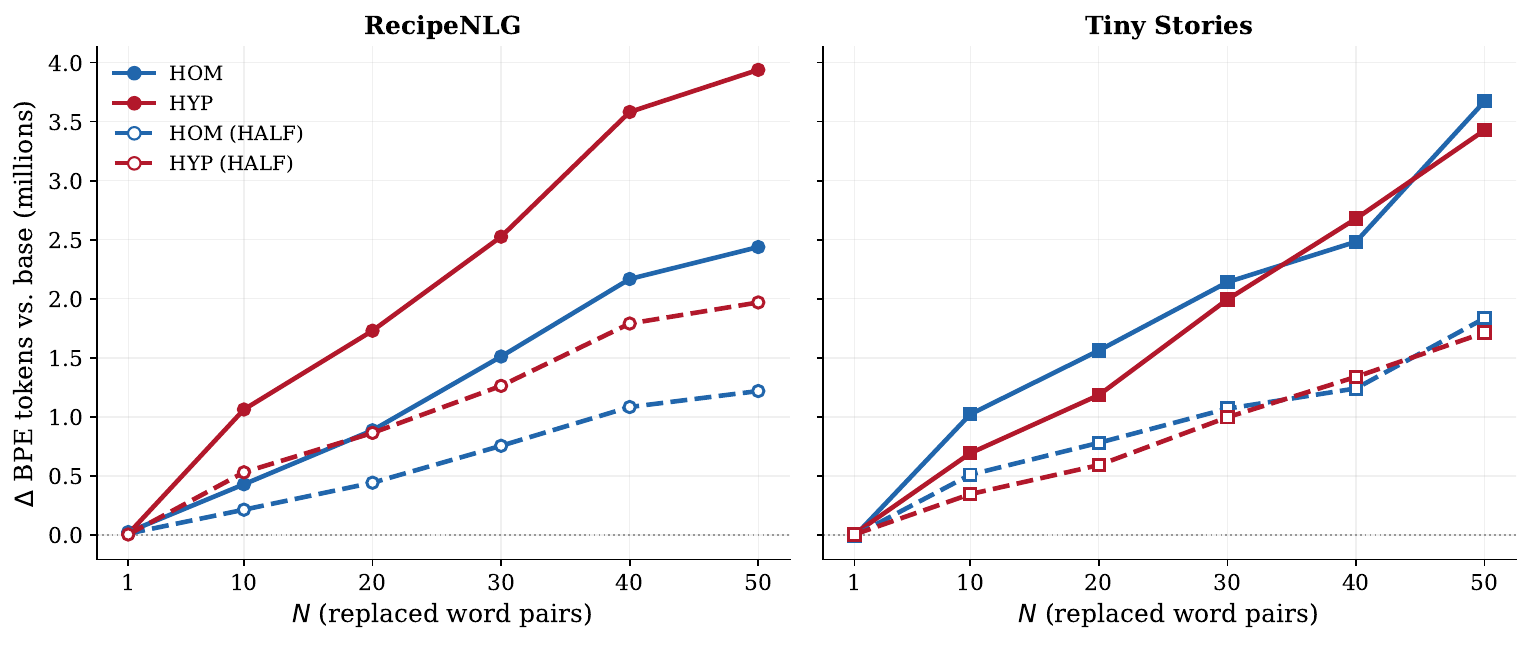}
 \caption{Absolute number of tokens in the models depending on the number of pseudoword types.}\label{fig:bpe}
\end{figure}

\section{Ambiguity Localization}\label{app:vne}
\begin{figure}[H]
 \centering
 \includegraphics[scale=0.5]{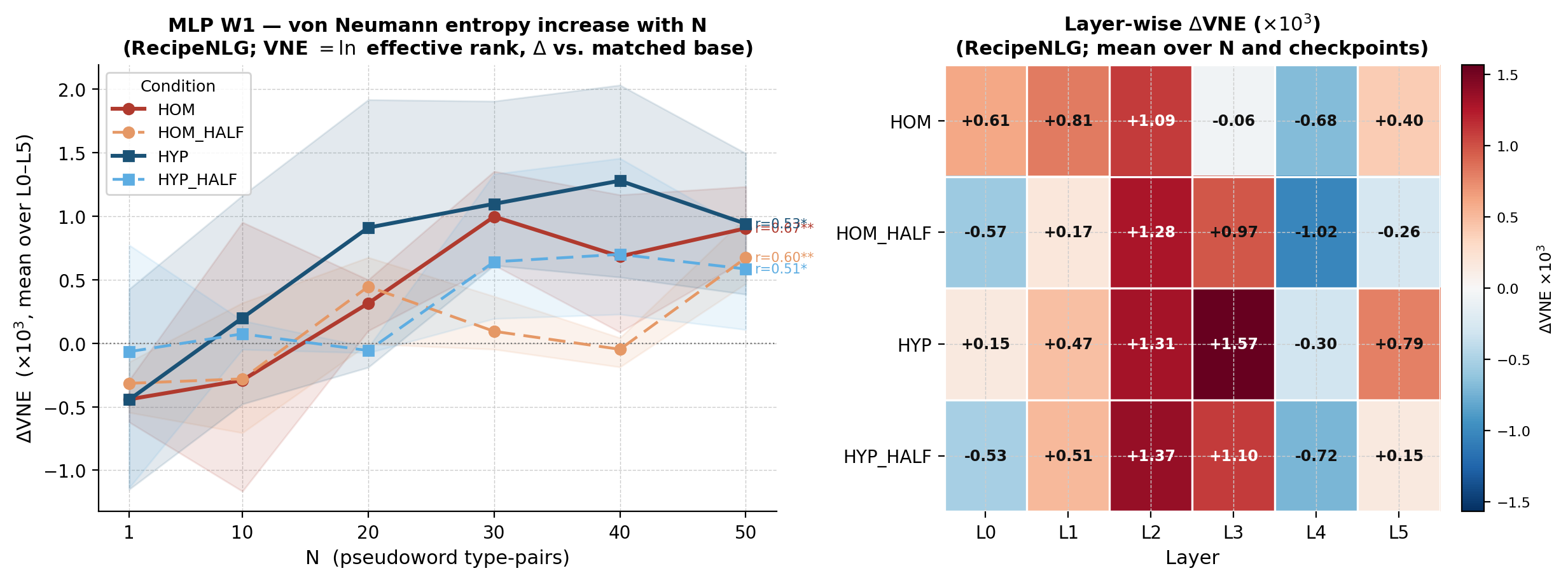}
 \caption{Von-Neumann-Entropy change in the models depending on increasing ambiguity or abstraction (left), and the layer-wise localization of that change (right).}\label{fig:vne}
\end{figure}
\section{Patterns for Disambiguation Probing}\label{app:dis}
\begin{table*}[h]
\small
\centering
\begin{tabular}{lll}
\hline
\textbf{ID} & \textbf{Neutral} & \textbf{Disambiguating} \\
\hline
1 & This is another $x$, like the other one. & This is another $x$, like the other $y$. \\
2 & Here we have that $x$, just an ordinary one. & Here we have that $x$, just an ordinary $y$. \\
3 & She likes that $x$, like all other things. & She likes that $x$, like all other $y$s. \\
4 & Look for my $x$, it looks like one of your things. & Look for my $x$, it looks like one of your $y$s. \\
5 & He talked about the $x$ after other topics. & He talked about the $x$ after other $y$s. \\
\hline
\end{tabular}
\caption{Matched patterns to test disambiguation. $x$ = pseudoword; $y$ = component word
(pluralized in frames 3--5). Each frame also generates a reference sentence
with $y$ as subject (\emph{$y$-neutral}) and a sanity-check sentence with
$y$ as subject and $x$ as comparison item (\emph{$y$-swapped}).}
\label{tab:frames}
\end{table*}
\begin{figure}[H]
\centering
 \includegraphics[scale=0.5]{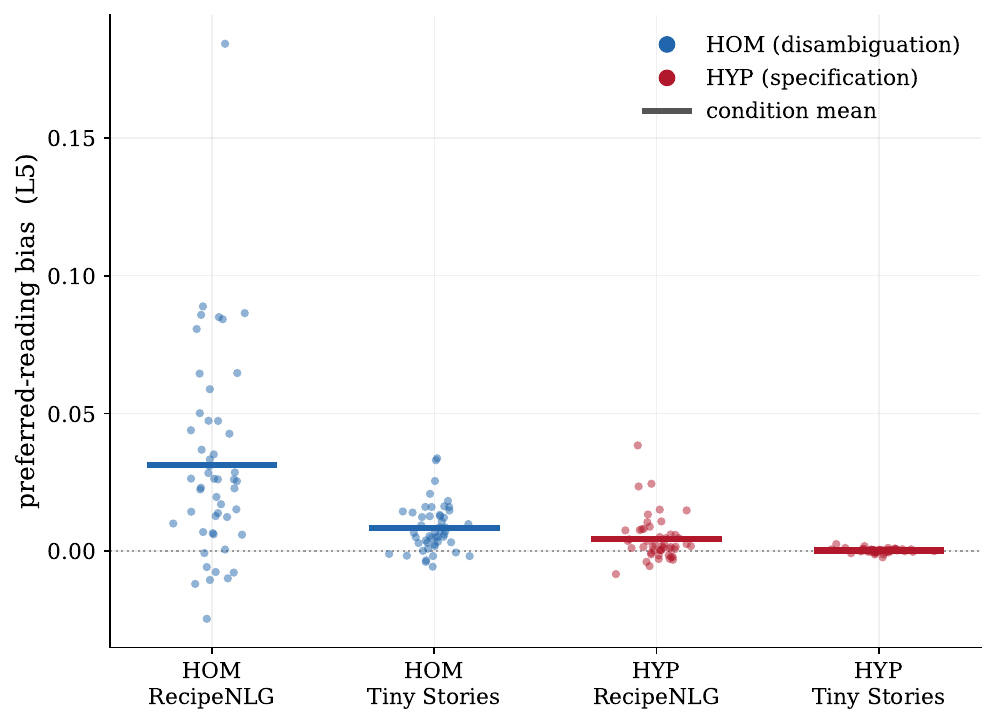}
 \caption{Sense bias for ambiguous and underspecified words. The dots show the differences in cosine similarity between the target word and the two senses, all in neutral contexts. A positive value indicates a bias towards the more frequent sense.}\label{fig:sense}
\end{figure}
\end{document}